\documentclass[a4j,10pt]{article}

\usepackage[dvipdfmx]{graphicx}     % 図はPDF形式
\usepackage{setspace}               % 行間調節用パッケージ
\usepackage{cite}                   % 引用スタイル
\usepackage{amsmath, amssymb,amsfonts, amsthm}  % 数学パッケージ
\usepackage{physics}                % 物理パッケージ(https://qiita.com/HelloRusk/items/ce9f49e9b3fc0344ae23)
\usepackage{lineno}
\usepackage{color}                  % 色付け用パッケージ
\usepackage{authblk}                % 著者の所属を記載するパッケージ

\begin{document}

\title{Model-Based Reinforcement Learning Exploits Passive Body Dynamics for High-Performance Biped Robot Locomotion
}

% 著者
\author[1]{Tomoya Kamimura}
\author[2]{Haruka Washiyama}
\author[2]{Akihito Sano}

% 所属
\affil[1]{\small Department of Mechanical Science and Bioengineering, Graduate School of Engineering and Science, The University of Osaka, Toyonaka, Japan. {\tt kamimura.tomoya.es@osaka-u.ac.jp}}
\affil[2]{\small Department of Electrical and Mechanical Engineering, Nagoya Institute of Technology, Aichi, Japan.}

\date{} % 日付を空欄にする

\setlength{\baselineskip}{4.4mm}	% 行間の設定
\maketitle
\setstretch{1.2} % ページ全体の行間を設定

% =========================
% 目次を自動生成するには以下をコメント解除
% \tableofcontents
% \newpage
% =========================

\begin{abstract}
  Embodiment is a significant keyword in recent machine learning fields. This study focused on the passive nature of the body of a biped robot to generate walking and running locomotion using model-based deep reinforcement learning. We constructed two models in a simulator, one with passive elements (e.g., springs) and the other, which is similar to general humanoids, without passive elements. The training of the model with passive elements was highly affected by the attractor of the system. This lead that although the trajectories quickly converged to limit cycles, it took a long time to obtain large rewards. However, thanks to the attractor-driven learning, the acquired locomotion was robust and energy-efficient. The results revealed that robots with passive elements could efficiently acquire high-performance locomotion by utilizing stable limit cycles generated through dynamic interaction between the body and ground. This study demonstrates the importance of implementing passive properties in the body for future embodied AI.
\end{abstract}

% 以下，本文
\section{Introduction}
The significance of the robot body has recently attracted the attention of machine learning researchers, especially in legged robots \cite{Hwangbo2019-bx,Duan2022-ly,Wang2023-yx, Hoeller2024-jr, Radosavovic2024-tm,Haarnoja2024-at}.
However, it is unclear how to properly construct the body of a robot and design a controller because of the complexity of legged robot dynamics.
For example, they are hybrid systems in which the governing equations change depending on the stance condition of the legs.
Furthermore, because running includes flight phases in which all legs are in the air, the robot is under-actuated when running.

In contrast, humans and animals produce adaptive locomotion by skillfully using their complex and redundant musculoskeletal systems.
It has been said that their locomotion is a dynamic phenomenon generated through the dynamic interaction between the neurons, body, and environment (e.g. the ground) \cite{Aoi2017-kj,Fukuhara2021-nd,Owaki2021-yq,Ijspeert2023-qc}.
For example, attractors, such as stable limit cycles, are generated as a result of the dynamic interaction \cite{McGeer1990-ab, Collins2005-kf,Srinivasan2006-uw, Geyer2006-ms, Adachi2020-oj,Kamimura2021-wv,Kamimura2023-ml}.
The locomotion that uses such an attractor is natural and energy efficient \cite{Pfeifer2007-yl,Fukuhara2022-pm,Fukuhara2022-xw,Badri-Sprowitz2022-fz}.

It is essential to understand the mechanisms under such skillful locomotion of humans and animals using machine learning approaches, and to use these methods for efficient control.
In conventional robot learning, locomotion is constructed from scratch, and the cost of control tends to be too high.
While the learning process takes a long time, the control forces or torques are too high, or the energy efficiency is low.
In contrast, if attractors generated through the system's dynamics are used, exploration is efficient, the control has a low dimensionality, and the system is more energy efficient because its dynamics determine locomotion.
Many studies have been conducted to acquire control laws using reinforcement learning for embodied robots \cite{Tani1996-yi}.
Furthermore, researchers have been investigating the efficient acquisition of robust animal-like locomotion by combining machine learning and neuron models such as central pattern generators (CPGs) \cite{Deshpande2023-cr,Li2024-xu}.
However, the effect of dynamic interaction between the body and ground, which plays a significant role in locomotion, on machine learning has not been sufficiently explored.
It is unclear what physical features are necessary and what hardware design should be used for the robot when performing such learning.

In contrast, some researchers recently focused on the effect of biomimetic body design~\cite{Wang2019-ni,Weng2021-lc,Ogum2024-rx}.
Some of them focus primarily on the passive characteristics of the body \cite{Yamaguchi2021-yk, Koseki2023-px,Bjelonic2023-pj}.
We have also been developing biped walking and running robots based on biomimetic musculoskeletal systems with passive properties, such as springs and highly backdrivable actuators, based on the principles of passive walking \cite{McGeer1990-ab,Collins2005-kf,Ikemata2006-mi} and the principle of bouncing rod dynamics \cite{Miyamoto2010-ny}.
The robot can walk and run without incorporating sophisticated control schemes such as model predictive controllers because of the passive nature of the body \cite{Sakurai2024-rt}.

The dynamic characteristics of passive bodies are another critical aspect when using machine learning.
Understanding the dynamic mechanisms that make this possible is expected to substantially contribute to the development of embodied AI in the future.
The imitation learning of human locomotion, or reinforcement learning of manipulation and locomotion tasks have been enhanced using musculoskeletal models based on biomechanical findings \cite{Anand2019-nj,Wochner2022-oe}.

This study investigates how the embodiment, especially the passive dynamics of the body, is used in learning by focusing on discovering limit cycles through reinforcement learning.
We clarify how the embodiment is used in learning without prior information using a robot with slightly different body designs from humans rather than imitation learning.
We construct two biped robot models in a dynamics simulator, one with a passive nature with high backdrivability and elastic elements and the other consisting of typical servomotors with low backdrivability.
These robot models are trained with identical rewards and hyperparameters in deep reinforcement learning based on the world model \cite{Ha2018-hg,Hafner2020-zv,Wu2022-in}, which can predict the dynamics of the system and use it for control.
We investigate which of them converges faster and whether the learning results are robust, discussing the advantages of the passive nature of the robot.

\section{Methods}
\subsection{Robot models}

\begin{figure}
        \centering
        \includegraphics[scale = 1.2]{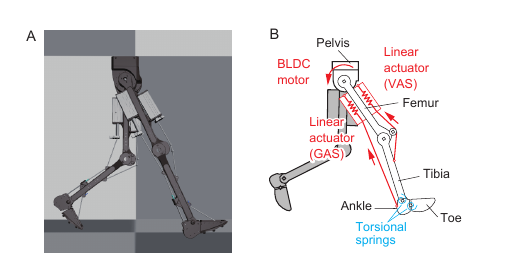}
        \caption{Biped robot and model. (A) Lower body model based on musculoskeletal structure robot in MuJoCo. (B) Schematics of the passive model.}
        \label{fig:robot}
\end{figure}

We have been developing a biped robot based on the passive nature of dynamics~\cite{Sakurai2024-rt}.
In this study, to elucidate how the passive dynamics of the musculoskeletal robot contribute to learning, we proposed a simplified robot model (Fig.~\ref{fig:robot}).
Specifically, this model was developed by extracting only the lower body from a previously developed musculoskeletal robot using the dynamics simulator MuJoCo.
The leg length and body mass are $0.78$~[m] and $6.38$~[kg], respectively. The detailed length, mass, and inertial parameters are shown in Table~\ref{table:passive_model_mass}.

\begin{table}
  \centering
  \caption{Length (distance between joints), mass and inertia (around each center of mass) of both passive and torque model}
  \label{table:passive_model_mass}
  \begin{tabular}{c|c|c|c}
    %\hline
    {Element} & Length [m] & {Mass} [kg] & {Inertia} [kgm$^2$] \\
    \hline
    Pelvis & N/A    & $2.77$                & $\infty$ \\
    Femur  & 0.345  & $1.4$                 & $1.1\times 10^{-2}$ \\
    Tibia  & 0.447  & $2.2 \times 10^{-1}$  & $3.0\times 10^{-3}$ \\
    Ankle  & 0.036  & $7.6 \times 10^{-2}$  & $6.1\times 10^{-5}$ \\
    Toe    & 0.072  & $8.3 \times 10^{-2}$  & $9.0\times 10^{-5}$ \\
  \end{tabular}
\end{table}

A model with passive elements uses two linear actuators (pneumatic actuators in real robots) to replicate some of the muscles of the human lower limb.
Specifically, linear actuators that simulate the vastus (VAS) and gastrocnemius (GAS) muscles are attached to the femurs to replicate the function of these muscles.
The elastic characteristics of each muscle is reproduced by attaching springs in series to wires built with tendon parts.
Two passive rotational springs are introduced at the ankle and foot joints to model the effects of the anterior and sole tibialis, and foot muscles, respectively.
The hip joint is driven by a quasi-direct drive BLDC motor (RMD-X8, MyActuator) with a reduction ratio of 6:1, same as~\cite{Sakurai2024-rt}.
We represented the inertia of the motor rotor by adding a motor rotor inertia of $1.22\times 10^{-2}$~[kgm$^2$], which includes the effect of the gear reduction ratio, to the hip joints.
Totally the model has six actuators, three in each leg.
The model is constrained on the sagittal plane; the pelvis link does not involve rotation in pitch direction.
We call this the passive model.
The parameters of each elastic element are listed in Table~\ref{table:passive_model}.
Although actuators are active elements, we focused on the passive characteristics of actuators in this study because we use actuators with high backdrivability, and their drive shafts can be moved by external forces.

\begin{table}
  \centering
  \caption{Spring and dumper parameters of passive model}
  \label{table:passive_model}
  \begin{tabular}{c|c|c}
	% \hline
	{Element} & {Spring constants} & {Damping coefficients} \\
	\hline
	VAS & $8000$ [N/m] & $100$ [Ns/m] \\
	GAS & $3000$ [N/m] & $100$ [Ns/m]\\
	Ankle & $100$ [Nm/rad] & $0.1$ [Nms/rad]\\
	Foot & $10$ [Nm/rad] & $0.2$ [Nms/rad]\\
	% \hline
	\end{tabular}
\end{table}

For comparison, we built another model that reproduces a general humanoid driven by servomotors by removing the pneumatic actuator and spring-wire system, and replacing it with high-reduction ratio motors (Dynamixel PM54–060-S250, reduction ratio 251:1) to each joint of the passive model.
To eliminate passive behavior of actuators, we reproduced low backdrivability caused by the high-reduction ratio motor by adding a motor rotor inertia of $22.83$~[kgm$^2$] to the joints.
Since the robot has six servo motors, the number of actuators is identical to that of the passive model.
Furthermore, we added high viscosity to the joint parts.
We call this the torque model as in the previous studies~\cite{Anand2019-nj,Wochner2022-oe}.

\subsection{Deep reinforcement learning}

\begin{figure}
  \centering
  \includegraphics[scale = 0.8]{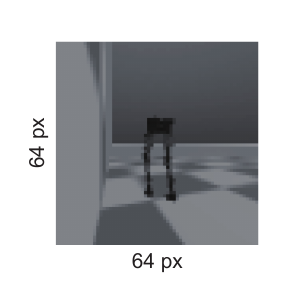}
  \caption{Monochrome image of the robot taken from behind, whose size is 64 $\times$ 64 px.}
  \label{fig:state_figure}
\end{figure}

In this study, controllers were obtained via model-based reinforcement learning to achieve locomotion that uses the interaction between the robot body and ground.
For consistency between the states of the two models, we defined the states as a camera image of the robot taken from a left diagonal rear angle, as illustrated in Fig.~\ref{fig:state_figure}.
As a deep reinforcement learning method, we used DreamerV2 \cite{Hafner2020-zv,Wu2022-in}, one of the world-model-based reinforcement learning methods, to efficiently reduce the dimension of images, capture features, and use time series data.

The action space for the motor torque and the force of the linear actuator was defined with discrete values.
In the two robot models, the actuators of both legs are always set to move in the opposite phase to reduce the number of actions.
We defined the action in the passive model as $a_{\rm p}=[f_{\mathrm{vas}} \ f_{\mathrm{gas}} \ \tau]$, where $f_{\mathrm{vas}}$, $f_{\mathrm{gas}}$, and $\tau$ are the forces of the linear actuators of the vastus and gastrocnemius muscles attached to the front and back of each leg, respectively, and the torque of the hip motor.
In the torque model, the action in the torque model is $a_{\rm t} = [\tau_{\mathrm{hip}} \ \tau_{\mathrm{knee}} \ \tau_{\mathrm{ankle}}]$, where $\tau_{\mathrm{hip}}$, $\tau_{\mathrm{knee}}$, and $\tau_{\mathrm{ankle}}$ are the torques of the motors in the hip, knee, and ankle joints, respectively.
Regardless of the robot model, the same number of candidate numbers of actions were used to make reasonable comparisons.
The action space of the passive and torque models is presented in Tables~\ref{table:action_passive} and~\ref{table:action_servo}, respectively.
The output torque of the motors is denoted as output shaft torque, which includes the effect of the reduction gear ratio.
Torque and force values of the passive model are determined according to real robot specifications.
Motor torque values of the torque model are adjusted heuristically based on the actual specifications of the servo motor (Dynamixel PM54–060) to make the robot walk and run.

\begin{table}
	\caption{Actions in passive model}
	\label{table:action_passive}
	\centering
	\begin{tabular}{ll}
		\hline
		Output & Value \\
		\hline \hline
		$f_{\mathrm{vas}}$~[N] & $[+400, 0, -400]$	\\
		$f_{\mathrm{gas}}$~[N] & $[0, -400]$	\\
		$\tau$~[Nm] & $[+24, +18, -18, -24]$	\\
		\hline
	\end{tabular}
\end{table}

\begin{table}
	\caption{Actions in torque model}
	\label{table:action_servo}
	\centering
	\begin{tabular}{ll}
		\hline
		Output & Value \\
		\hline \hline
		$\tau_{\mathrm{hip}}$~[Nm] & $[+753, +502, -502, -753]$	\\
		$\tau_{\mathrm{knee}}$~[Nm] & $[+502, 0, -251]$	\\
		$\tau_{\mathrm{ankle}}$~[Nm] & $[+251, 0]$	\\
		\hline
	\end{tabular}
\end{table}

The reward $r$ was set as follows.
\begin{align}
	r =  w_{\mathrm{v}}r_{\mathrm{v}} + w_{\mathrm{h}} r_{\mathrm{h}}
\end{align}
where
\begin{align}
    r_{\mathrm{v}} &=
  \begin{cases}
    \dot{x} - v_0   & (\dot{x} \leq v_{\mathrm{d}} ) \\
		2v_{\mathrm{d}} - \dot{x} - v_0  & (\dot{x} > v_{\mathrm{d}})
  \end{cases}
  \label{eq:reward_v}\\
    r_{\mathrm{h}} &=
  \begin{cases}
    -1    & (z < z_0) \\
    0     & (z \geq z_0)
  \end{cases}
  \label{eq:reward_h}
\end{align}
and $r_{\mathrm{v}}$ and $r_{\mathrm{h}}$ are the rewards from the horizontal velocity of the pelvic link $\dot{x}$ and to the height of the pelvic link $z$, respectively, $w_{\mathrm{h}}$ and $w_{\mathrm{v}}$ are the weight coefficients and $v_{\mathrm{d}}$ is the target velocity.
$r_{\mathrm{v}}$ is the reward for the forward speed, which is maximized at the target speed.
It is also penalized below a threshold $v_0 = 0.2$~[m/s] to prevent standstill.
$r_\mathrm{h}$ is a reward for preventing falls and a penalty if the height of the pelvic link of the robot falls below a threshold $z_0 = 0.7$~[m].
In this study, $w_{\mathrm{h}} = 1$ and $w_{\mathrm{v}} = 1 / (v_{\mathrm{d}} - 0.2)$ were set so that the maximum reward per step was $1$.
We set the target speeds to $1.5$~[m/s] and $2.5$~[m/s], expecting to obtain walking and running, respectively.

In this study, we used the identical learning model, hyperparameters, and rewards for both robot models.
A learning step was set to $50$~[ms], and the $500$~steps were defined as one episode.
$40$ episodes with random actions are performed before the start of reinforcement learning to train the world model.

\subsection{Evaluation of reinforcement learning}
We compared the learning curves to compare the learning efficiency between the two models.
Furthermore, we investigated how periodic trajectories are generated through learning processes.
Because the state variables are high-dimensional, the time profiles of the six joint angles (three joints in each leg) are reduced in dimension using Uniform Manifold Approximation and Projection (UMAP) \cite{McInnes2018-gc} for visualization and projected into a two-dimensional space.

\subsection{Evaluation of obtained locomotion}
The obtained locomotion is compared with the joint angle kinematics of human walking~\cite{Fukuchi2018-fj} and running~\cite{Fukuchi2017-dr}.
Although the passive model has a musculoskeletal structure similar to that of a human, the hip joint involves a servomotor.
Therefore, it is assumed that the optimized locomotion of such a hybrid design robot obtained by reinforcement learning without prior information is not identical to that of a human.

Furthermore, we compared the energy efficiency to evaluate the obtained locomotion performance.
Because the two robot models have identical morphology and mass, it was sufficient to investigate the energy consumption.
However, because it is impossible to determine the exact energy consumption (including friction) on the simulator, we instead calculated the amount of work by the actuators over a given period.

Furthermore, to compare the robustness of the passive model and the torque model, we used the inference model obtained from the learning results as is and made the robot walk and run on various slopes with slope angles $\alpha \in \{ -5, -3, 3 \}$~[deg], where a positive angle indicates an ascending slope.

\section{Results}
\subsection{Learning on level ground}
\begin{figure}
  \centering
  \includegraphics[scale = 1.2]{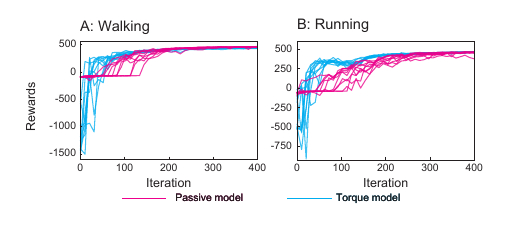}
  \caption{Learning curves of passive model and torque model in 10 trials. (A) Walking at 1.5~[m/s]. (B) Running at 2.5~[m/s].}
  \label{fig:rl_results}
\end{figure}

As a result of learning, for each robot model, walking and running were acquired for $v_{\rm d} = 1.5$~[m/s] and $2.5$~[m/s], respectively, although the action space is discrete and the number of action candidates is highly limited.
The rewards for each robot model and target speed for $10$ trials are depicted in Fig.~\ref{fig:rl_results}.
Because the passive model is more prone to falls than the torque model because of its backdrivability, the rewards at the beginning of the training are lower than those of the torque model.
Moreover, the rewards converged more slowly in the passive model, and the learning curve was not as steep as that of the torque model.
Consequently, in both robot models, the rewards converged almost to the maximum possible value (about $400$).
Note that the torque model sometimes exhibited backward motion in the early stages of learning, resulting in very small rewards (i.e., large absolute values), which was not observed in the passive model.

\begin{figure}
    \centering
    \includegraphics[scale = 1.2]{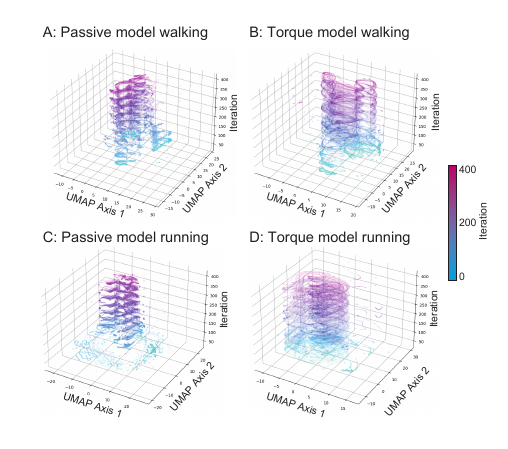}
    \caption{Convergence of trajectories with learning process. Two-dimensional projection of joint angles using UMAP. (A) Passive model walking, (B) torque model walking, (C) passive model running, and (D) torque model running.}
    \label{fig:limit_cycles}
\end{figure}

Figure~\ref{fig:limit_cycles} illustrates the evolution of the trajectory in the two-dimensional space reduced by UMAP as reinforcement learning progresses. Even if the trajectory is periodic, the structure in reduced-dimension space does not always appear as a closed orbit. With the passive model, the motion converges to a particular structure, which indicates a limit cycle, early in the iteration for both walking (Fig.~\ref{fig:limit_cycles}A) and running (Fig.~\ref{fig:limit_cycles}C). In contrast, with the torque model, there is no convergence to a particular structure for both walking (Fig.~\ref{fig:limit_cycles}B) and running (Fig.~\ref{fig:limit_cycles}D). The final results illustrate low density of plots, which indicate that the trajectory were not periodic.

\begin{figure}
  \centering
  \includegraphics[scale = 1.2]{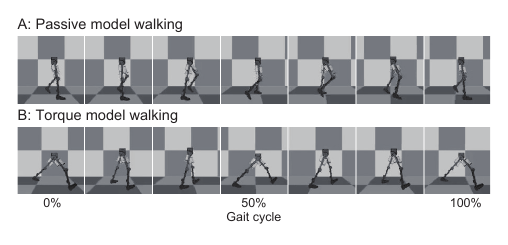}
  \caption{Snapshots of typical walking with $v_{\rm d} = 1.5$~[m/s] by (A) passive model and (B) torque model.}
  \label{fig:snapshots_walk}
\end{figure}

\begin{figure}
  \centering
  \includegraphics[scale = 1.2]{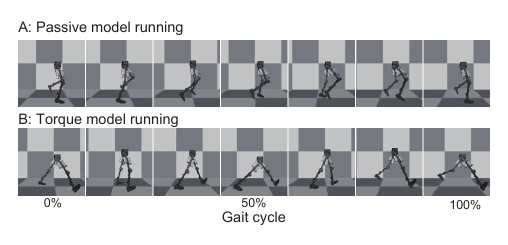}
  \caption{Snapshots of typical running with $v_{\rm d} = 2.5$~[m/s] by (A) passive model and (B) torque model.}
  \label{fig:snapshots_run}
\end{figure}

Although rewards converged to similar values for the two robot models at both speeds, the resulting locomotion at the end of training was qualitatively different for each robot.
Typical locomotion obtained for target speeds $v_{\rm d} = 1.5$~[m/s] and $2.5$~[m/s] are depicted in Figs.~\ref{fig:snapshots_walk} and~\ref{fig:snapshots_run}, respectively.
Regardless of the target speed, the passive model produced soft and bending joint motions, especially while running, whereas the torque model produced stiff joint motions.
Furthermore, regardless of the moving speed, the torque model exhibited larger step lengths than the passive model.

\subsection{Evaluation of obtained locomotion}
\begin{figure}
  \centering
  \includegraphics[scale = 1.2]{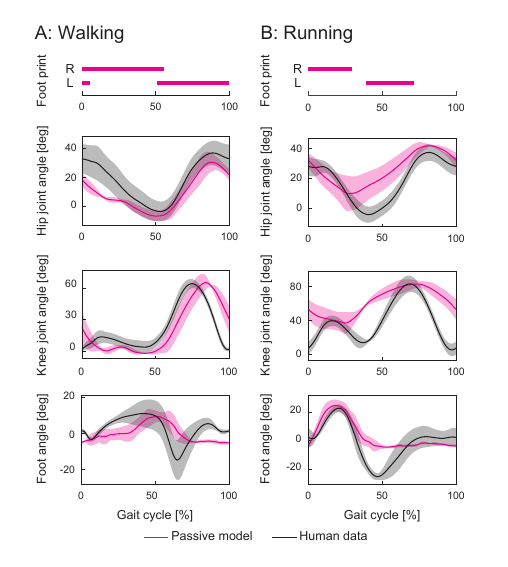}
  \caption{Footprint diagrams (model only) and time profiles of joint angles in typical obtained locomotion (Figs.~\ref{fig:snapshots_walk} and~\ref{fig:snapshots_run}) of passive model (red) compared with human data (black). (A) Walking and (B) running. ``R'' and ``L'' indicate right and left foot, respectively. Solid lines and shaded areas represent the mean and standard deviation of 5 strides.}
  \label{fig:time_profiles}
\end{figure}

Footprint diagrams and the time profiles of the joint angles in the right leg of the typical walking and running obtained in the passive model are depicted in Fig.~\ref{fig:time_profiles}, where one gait cycle is determined as a period from the landing of the right leg till the landing of the same leg.
The foot angle is defined for the robot model as the sum of the ankle and foot joint angles to compare with human ankle joint data because the fluctuations of ankle foot joint angles were much smaller and larger than those of human data, respectively.
As in human locomotion, the obtained walking and running involved a double support phase and a flight phase, respectively. Note that the passive model's left-right foot motion during walking is symmetric, while the foot motion during running is asymmetric.
Compared with human data, the time profiles of each joint in the passive model were qualitatively similar to those of humans in some cases.
Specifically, in both walking and running, the hip joints exhibit qualitatively similar kinematics in the robot model and humans.
In contrast, the motion of the knee joint when running was qualitatively different, with fewer peaks and smaller magnitudes in the model.
In both walking and running, the foot angle took small negative values (plantar flexion) around the mid-gait cycle, which differed from human locomotion.
Furthermore, the gait cycle periods differed significantly between the model and humans, with 0.37 and 0.9~[s] for walking and 0.29 and 0.7~[s] for running, respectively.

We also calculated the mean and standard deviation for the amount of energy exerted by the actuators in each model during one episode (for 25~[s]) in the 10 learning results.
In walking, the average and standard deviation of energy consumption of the passive and torque models were $17.0 \pm 4.7$ and $63.6 \pm 20.0$~[kJ].
In running, the energy consumption of the passive and torque models was $16.5 \pm 2.8$ and $154.9 \pm 6.5$~[kJ].

\begin{figure}
  \centering
  \includegraphics[scale = 1.2]{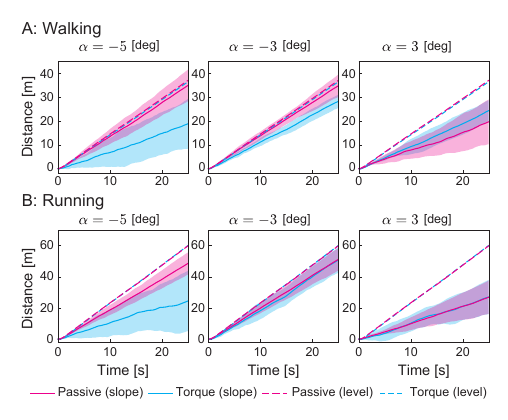}
  \caption{Horizontal position on slopes (solid) and level ground (dashed) with passive model (red) and torque model (blue) in (A) walking and (B) running. Lines and shaded areas represent the mean and standard deviation of 10 trials, respectively.}
  \label{fig:distance_slopes}
\end{figure}

Walking and running tests were performed on slopes with various inclinations $\alpha$ for each robot model to investigate the robustness of the reinforcement learning results on level ground.
Figure~\ref{fig:distance_slopes} illustrates the time profiles of the pelvic horizontal position during locomotion on slopes by the inference from the 10 learning results on level ground for $v_{\rm d}=1.5$ and $2.5$~[m/s].
The averaged time profiles for each target velocity on level ground using inference from 10 learning results are also depicted for comparison.
On descending slopes ($\alpha = -5$ and $-3$~[deg]), the passive model walked and ran similarly as on level ground. In contrast, the torque model sometimes went backward and could not move forward as much as on level ground, which resulted in small averages and large standard deviations in traveling distances.
This result indicates that the passive model has higher robustness to the torque model.
In contrast, for both walking and running, on the ascending slope ($\alpha = 3$~[deg]), both the robot models could not move forward as much as on level ground.

\section{Discussion}

\subsection{Effect of passive nature of the body on learning}
In this study, using identical deep reinforcement learning (DreamerV2) and reward, we performed reinforcement learning for biped robot models with and without the passive nature of the body.
As shown in Fig.~\ref{fig:rl_results}, the passive model experienced longer periods of low reward than the torque model.
The slope of the learning curve was smaller, and it took longer for the reward to converge, although the rewards saturated at about the same value for both robot models consequently.
However, the convergence of trajectories were qualitatively different.
In the passive model, the trajectories were entrained to a specific limit cycle early in the learning process (Fig.~\ref{fig:limit_cycles}A and C).
In contrast, for the torque model, the obtained trajectories did not converge to a specific periodic trajectory (Fig.~\ref{fig:limit_cycles}B and D).

These results could be attributed to the training of the passive model being strongly affected by the attractors of the system.
Specifically, in the early stages of training, the passive model tended to converge to a stable fixed point corresponding to falling, which led to small rewards.
Subsequently, the model became entrained to stable limit cycles corresponding to walking or running, and gained large rewards.
Note that the obtained limit cycles do not necessarily increase the rewards, which are artificially designed.
Because the limit cycles of the passive model gradually changed to increase the given reward (see attached video), it took a long time for the rewards to converge.
In contrast, the torque model progressed rapidly in training by generating locomotion that yielded large rewards by using large torques.

Although the torque model showed better learning curves than the passive model in this study, the difference in learning speed would depend on the reward.
It is expected that the learning speed could change if rewards suitable for the passive model, such as those utilizing attractors, are used rather than the rewards used in this study, such as speed and height.

Moreover, the obtained locomotion was partly qualitatively different from that of a human, as depicted in Fig.~\ref{fig:time_profiles}. The limit cycle varies depending on the robot's body design.
Although this might be caused by the difference between the natural human cost function, which is unknown, and our reward design, it can be suggested that robots must discover the optimal motion for their bodies distinct from the human body by reinforcement learning rather than imitation learning.

\subsection{Effect of passive nature of the body on performance}

The world model, which can predict the system's behavior, enabled locomotion considering the dynamic characteristics of each model.
Consequently, the passive model obtained movements in which the joints bent softly (Figs.~\ref{fig:snapshots_walk}A and \ref{fig:snapshots_run}A), while the torque model obtained movements in which the joints were stiff and rigid (Figs.~\ref{fig:snapshots_walk}B and \ref{fig:snapshots_run}B).
In the passive model, the stance leg was stretched like a stick during walking (Fig.~\ref{fig:snapshots_walk}A), while the joints were bent during stance phases in running (Fig.~\ref{fig:snapshots_run}A).
Such characteristics are also observed in human locomotion.
In walking, the knee is extended to behave like an inverted pendulum and save energy, while when running, the knee is bent to store energy \cite{Geyer2006-ms}.
The passive model acquired more energy-efficient locomotion than the torque model because it uses the passive nature of the body in a reasonable way, similar to human locomotion.

It was also confirmed that the movement of the passive model was robust.
Walking and running tests on a slope were performed using inference models from reinforcement learning on level ground.
On the descending slopes, while the passive model could move forward similarly as on level ground (Fig.~\ref{fig:distance_slopes}), the torque model sometimes involved backward movements.
Because the torque model has joints with low backdrivability and no passive elements, locomotion could result in inappropriate behavior by performing the same locomotion on the slope as on level ground.
In contrast, because the passive model has elastic elements and joints with high backdrivability, the robot body absorbs the differences in the terrain between the slope and level ground, and the interaction between the body and ground is assumed to have resulted in appropriate locomotion.
Moreover, the entrainment to the limit cycle also contributed to the robustness of the passive model.
Both robot models exhibited low robustness on the ascending slope. The torque model occasionally exhibited a longer average traveling distance than the passive model. This difference is assumed to be because the passive model uses the principles of passive walker~\cite{Ikemata2006-mi} and is suitable for a descending slope. In contrast, the torque model actively generates its behavior using their motors to inject the energy for ascending the slope.

Furthermore, because the joint backdrivability of the passive model is low, the robot model tends to fall easily, which is an undesirable behavior in traditional robotics.
However, in this study, in which the world model was used, it is considered to be an advantage that the robot model fell over many times at the beginning of the training.
The world model experienced various types of falls in the early phase of the training and can observe various types of dynamic behaviors of the robot body and the ground.
This enabled the construction of a world model that can represent a wider variety of input-output relationships and reproduce situations in which the robot falls over or fails to move forward.
Reinforcement learning based on such a world model might have produced a more robust controller.

\section{Conclusion}
Based on the idea of embodiment, this study demonstrated that the dynamic characteristics of the body have a significant contribution to machine learning.
Although a robot with passive elements required a long time to train, it acquired robust and energy-efficient locomotion with reinforcement learning by finding appropriate limit cycles depending on the robot body design because of the passive nature of the musculoskeletal body.
Furthermore, although high backdrivability leads to a tendency to fall, such experience may have provided diversity to the world model and led to a more robust control system.
These results indicate that the combination of a robot with the dynamic feature that involves limit cycles and the world model can efficiently learn high-performance locomotion.
In the future development of artificial intelligence with a body, the passive properties of the robot body, such as high backdrivability and elastic elements, and model-based reinforcement learning, which can understand and use the dynamics of the target, will be suitable for realizing sophisticated locomotion.

\subsection{Limitations and future works}

This study used simulation to compare robots with identical physical parameters but with different actuation systems.
However, in the passive model, qualitatively different passive elements were coupled. Specifically, the passive model involves completely passive elements such as springs, as well as the passive nature of actuators caused by high backdrivability. In future research, we need to propose an improved model in which these elements are decoupled to investigate how each element affects the learning results.
We also would like to conduct verifications using physical robots and confirm that the gap between simulation and real robots can be bridged more easily because of the robustness of the learning results.

Moreover, the results suggested that model-based reinforcement learning is a useful method for learning embodied locomotion. However, we must still clarify how the model affects the learning process and results, comparing them to model-free reinforcement learning.

Furthermore, in this study, we restricted learning locomotion only on level ground and a uniform planar slope. Future research should perform learning for various purposes, such as moving at various speeds, traveling on uneven terrain, and switching between walking and running, to apply the learning to real-world applications.

From a physiological point of view, the neural system (e.g., CPGs) significantly contributes to human and animal locomotion. We would like to construct a framework for more efficient learning by separating the roles of the brain and the nervous system.

\section*{Acknowledgement}
This work was supported in part by JSPS KAKENHI Grant Numbers JP21K14104, JP23K22716, and JP24K17237.
The manuscript partially used ChatGPT-4.1 for language improvement.

% \clearpage
\small
\bibliographystyle{IEEEtran}
\bibliography{2025_ICRA_Washiyama}
\normalsize

\end{document}